# Music Recommendation Based on Facial Emotion Recognition

Rajesh B[1] , Keerthana V[1],  Narayana Darapaneni[2], Anwesh Reddy P[3,*]

[1] PES University, Bangalore, Karnataka, 560050, India
[2]Northwestern University, Evanston, IL 60208, United States
[3]Great Learning, Hyderabad, Telangana, 500089, India

## Abstract

INTRODUCTION: Music provides an incredible avenue for individuals to express their thoughts and emotions, while also serving as a delightful mode of entertainment for enthusiasts and music lovers.
OBJECTIVES: This paper presents a comprehensive approach to enhancing the user experience through the integration of emotion recognition, music recommendation, and explainable AI using GRAD-CAM.
METHODS: The proposed methodology utilizes a ResNet50 model trained on the Facial Expression Recognition (FER) dataset, consisting of real images of individuals expressing various emotions.
RESULTS: The system achieves an accuracy of 82% in emotion classification. By leveraging GRAD-CAM, the model provides explanations for its predictions, allowing users to understand the reasoning behind the system's recommendations. The model is trained on both FER and real user datasets, which include labelled facial expressions, and real images of individuals expressing various emotions. The training process involves pre-processing the input images, extracting features through convolutional layers, reasoning with dense layers, and generating emotion predictions through the output layer
CONCLUSION: The proposed methodology, leveraging the Resnet50 model with ROI-based analysis and explainable AI techniques, offers a robust and interpretable solution for facial emotion detection paper.

## 1. Introduction

In the current digital age, the significance of user experience is pivotal across a wide range of applications.This paper seeks to enhance user experience by integrating three key components: emotion recognition, music recommendation, and explainable AI. The proposed system leverages deep learning techniques and GRAD-CAM for accurate emotion recognition and personalized music recommendations. Recent studies in the domain of music have revealed that music evokes strong emotional reactions in its listeners [1]. There is a significant correlation between musical preferences and personality traits as well as emotions. The areas of the brain responsible for emotions and mood control the various aspects of music, such as meter, timbre, rhythm, and pitch [2].

Elements such as sex, age [3], cultural background [4], mood, individual choices, and contextual factors [5] collectively influence an individual's emotional reaction to a specific piece of music, taking into account variables like the time of day or the setting. Notwithstanding these external influences, humans possess the ability to consistently classify music into categories such as happiness, sadness, excitement, or calmness.

Facial expressions possess considerable potential as indicators of an individual's psychological well-being, serving as the most fundamental and instinctive means of expressing emotions [6, 7, 2]. Despite the strong association, a majority of current music software lacks the capability to generate playlists aligned with diverse emotional states. Emotion recognition plays a vital role in understanding users' emotional states and tailoring content accordingly. By accurately detecting and classifying facial emotions, the system can adapt its functionality and provide content that resonates with users' emotional needs. This capability opens up new possibilities for interactive applications, such as personalized music recommendation systems.

Music recommendation systems have gained immense popularity in recent years, leveraging advanced algorithms to suggest songs based on users' preferences. However, integrating emotion recognition into music recommendation adds a new dimension to the system. By considering the detected emotions, the system can recommend songs that align with users' emotional states, creating a more meaningful and engaging music listening experience..

Explainable AI is another crucial aspect of the proposed system. Conventional machine learning models frequently lack interpretability, posing a challenge for users to comprehend the rationale behind their recommendations. By incorporating the GRAD-CAM technique, the system provides visual explanations for its predictions, enabling users to grasp the underlying features that contribute to the recommended content. This transparency fosters trust, understanding, and user engagement.

The primary objective of this study is to create an affordable music application that utilizes real-time video and Convolutional Neural Network (CNN) technology to automatically select songs based on the user's current mood state. The system aims to minimize resource consumption while incorporating an emotion module that analyzes the user's real-time video to evaluate their emotional state. Subsequently, it matches the identified mood with songs from a categorized collection and offers recommendations for a diverse range of songs. By alleviating the burden of manual song selection, this system has the potential to address the existing challenge in finding suitable songs.

## 2. Literature Survey

The literature survey encompasses a comprehensive analysis of existing research and studies related to emotion recognition, music recommendation systems, and explainable AI. This section highlights key findings, methodologies, and insights from previous works, laying the foundation for the proposed system.

Several studies have explored the use of different techniques and methodologies for music recommendation systems based on mood/emotions. In their study, Renu Taneja et al. [9] utilized Audio to retrieve audio features such as pace, beats, and RMSE. They then constructed clusters to represent various moods based on these extracted properties. Kee Moe Han et al. [8], on the other hand, employed the average emotions from a group of 15 individuals to determine the emotion of a song. They trained a classifier using this data and categorized the music signal's emotion by considering audio parameters such as pitch and timbre. Another research conducted by V. R. Ghule et al. [10] centered on the development of a music system that employed facial recognition technology for analyzing emotions.

In 2005, Wieczorkowska et al. conducted a study aiming to assist users in discovering music aligned with their moods. They employed the K-nearest neighbors (KNN) algorithm to classify a vast dataset of 327,683 songs into six distinct emotions, resulting in an overall accuracy of 37%. Similarly, in 2008, another study [12] utilized a regression method for Music Emotion Recognition (MER) and achieved accuracy rates of 64% for arousal and 59% for valence. Yading Song et al. [13] explored various facets of music for MER and utilized a labeled dataset of 2,904 songs categorized as "happy," "sad," "angry," or "relaxed." Support Vector Machines (SVM) were employed, with spectral characteristics exhibiting superior performance compared to other acoustic parameters.

In 1978, Ekman and Friesen [14] introduced Action Units (AU), which incorporated both permanent and transient facial traits for emotion recognition. The increasing popularity of employing Convolutional Neural Networks (CNNs) in emotion recognition can be attributed to the continuous advancements in methodologies. Lyrical analysis has also been utilized for music classification [15], [16]. However, relying exclusively on tokenized methods falls short in achieving accurate song categorization. Additionally, the presence of language barriers poses a limitation to classification within a single language, creating a distinct disadvantage in the overall process.

In the year 2020, T. Vijayakumar [17] introduced a research paper that concentrated on tackling inverse problems utilizing Convolutional Neural Networks (CNNs). The study initially employed CNN and later transitioned to direct inversion using a combination of Filtered Back Projection (FBP) and CNN, known as FBP-C. The approach utilized individual learning and a U-net architecture. The synthetic dataset used in the study consisted of 475 training images and 25 validation images. The backpropagation technique employed in the study produced satisfactory results.

In a recent study carried out in 2021, Sungheetha, Akey, and Rajesh Sharma [18] directed their attention towards image classification using Convolutional Neural Networks (CNN) for the early detection of Diabetic Retinopathy. The conventional methods employed for detecting Hard Exudates (HE) in retinopathy images, which are crucial for assessing diabetes severity, were found to be ineffective. To address this challenge, the study proposed the utilization of CNN to extract relevant features from deep networks, offering a viable solution. Deep learning architectures, including CNN, have demonstrated their effectiveness as powerful tools for image recognition, analysis, classification, and identification within the domain of medical imaging.

In the year 2021, a survey was conducted by Smys, S., Joy Iong Zong Chen, and Subarna Shakya [19] to investigate various architectures and design methodologies employed in neural networks. The study classified deep neural networks into three distinct types: hybrid architectures, generative architectures, and discriminative architectures. The hybrid architecture was presented by integrating Convolutional Neural Networks (CNN) with deep belief networks, while the discriminative architecture predominantly relied on CNN, featuring stacked pooling and convolution layers to construct a deep model. The survey provided insights into the diverse approaches and structures employed.

Emotion recognition has emerged as a highly investigated domain within the fields of computer vision and human-computer interaction. Researchers have explored various techniques, including facial expression analysis, physiological signals, and audio analysis. In the realm of music psychology, Swathi Swaminathan and E. Glenn Schellenberg [1] conducted research to shed light on the current state of emotion research, emphasizing the importance of comprehending emotions within music-related contexts. F. Abdat, C. Maaoui, and A. Pruski [2] directed their attention toward human-computer interaction and highlighted the significance of facial cues in emotion detection through facial expression recognition. These studies offer valuable insights into the theoretical foundations and practical applications of techniques utilized in emotion recognition.

Music recommendation systems have the objective of delivering personalized and pertinent music suggestions to users, taking into account their preferences, context, and emotional states. In the realm of mood classification from musical audio, Kyogu Lee and Minsu Cho [4] investigated the utilization of user group-dependent models, highlighting the importance of acknowledging user diversity and individual preferences in the realm of music recommendation. In a distinct perspective, Daniel Wolff, Tillman Weyde, and Andrew MacFarlane [5] concentrated on culture-aware music recommendation, recognizing the influence of cultural background on music preferences. Additionally, Mirim Lee and Jun-Dong Cho [15] developed a context-based social music recommendation service, underscoring the significance of contextual factors in augmenting music recommendation systems. These studies contribute to the comprehension of music recommendation techniques and the factors that impact user satisfaction and engagement.

Explainable AI has gained considerable attention to enhance transparency and interpretability in AI models. Peter Burkert, Felix Trier, Muhammad Zeshan Afzal, Andreas Dengel, and Marcus Liwicki [21] presented DeXpression, a deep convolutional neural network designed for expression recognition. This network integrates explainable AI techniques to provide visualizations of the specific image regions that contribute to the model's predictions. This study demonstrates the potential of explainable AI in enhancing the interpretability of deep learning models. The literature survey explores the existing research and developments in the field of explainable AI and its relevance to the proposed project.

By reviewing these studies and research papers, it is evident that emotion recognition, music recommendation, and explainable AI are active areas of research with various approaches and techniques. However, there is still a need to integrate these domains to enhance user experience. The proposed project aims to bridge these gaps and provide a comprehensive system that combines emotion recognition from facial cues, music recommendation based on emotions, and explainable AI using GRAD-CAM visualization..

# 3. Methodology

As emphasized in the literature review, Peter Burkert, Felix Trier, Muhammad Zeshan Afzal, Andreas Dengel, and Marcus Liwicki [21] presented DeXpression, a deep convolutional neural network designed for expression recognition. This network integrates explainable AI techniques to provide visualizations of the specific image regions that contribute to the model's predictions. Building upon the insights gained from the extensive literature survey, this paper now proceeds to outline the methodology employed to address the research gaps identified in the existing studies. The methodology section represents a fusion of established techniques from prior research and innovative approaches tailored to the specific objectives of our project.

By capitalizing on the strengths and limitations of previous methodologies, we have designed a refined framework that aims to advance the fields of facial emotion recognition and music recommendation. This comprehensive methodology encompasses various stages, including dataset acquisition, preprocessing, model selection, training, and evaluation. While also introducing novel strategies to tackle the unique challenges associated with facial emotion recognition and music recommendation.

## 3.1. Dataset Description

The proposed system utilizes a dataset consisting of facial images with labeled emotions for training and evaluation purposes. The dataset includes real images of different individuals encompassing a diverse range of emotions. The dataset may be augmented and pre-processed to enhance model performance and generalize well to real-world scenarios.

The dataset used for training the facial emotion recognition model consists of two components: the FER dataset and real images of different individuals. The dataset for facial expression recognition (FER) consists of categorized facial expressions, encompassing a range of emotions including anger, disgust, fear, happiness, sadness, surprise, and neutrality. This dataset serves as the foundation for training the deep learning model and enables it to learn patterns associated with various emotions. To enhance the diversity and generalization capability of the model, real images of different individuals are also captured and included in the training dataset.

In addition to the FER dataset, a music dataset is utilized for generating personalized music recommendations. The music dataset contains a diverse collection of music tracks from various genres and styles. This information serves as the basis for mapping the detected emotions to appropriate music tracks. The dataset for music can be acquired from diverse sources, such as online music platforms, curated databases, or personalized collections.

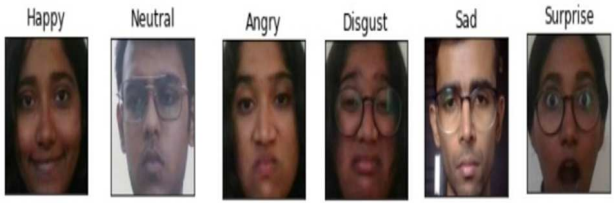

**Figure 1.** Sample images from the dataset.

By incorporating both the Real user image dataset and a comprehensive music dataset, the system can offer personalized music recommendations based on the user's real-time emotional state. The facial emotion recognition model trained on the FER dataset enables accurate emotion detection, while the music dataset provides a rich selection of tracks for mapping and playlist generation. Table 1 provides the classifications of moods and corresponding songs.

Table 1. Music Dataset

| Sr.No | Emotions | No of Songs |
|-------|----------|-------------|
| 1 | Happy | 20 |
| 2 | Sad | 30 |
| 3 | Angry | 20 |
| 4 | Surprise | 20 |
| 5 | Neutral | 20 |
| 6 | Disgust | 20 |
| 7 | Fear | 16 |

## 3.2. Dataset Description

Facial emotion recognition utilizes the ResNet50 model, a deep CNN architecture renowned for its exceptional performance in image classification assignments It consists of 50 layers, including convolutional layers, shortcut connections, and global average pooling. The input layer of the ResNet50 model takes in facial images as input. The convolutional layers extract meaningful features from the input images, capturing the facial expressions' key characteristics. The dense layers process the extracted features and learn the relationship between facial expressions and emotions. Finally, the output layer provides the predicted emotion based on the learned features.

The output of the dense layer in the network employs the softmax activation function, which enables the prediction of a multinomial probability distribution. This distribution is well-suited for multiclass classification tasks that involve more than two labels. In this particular project, which involves classifying emotions into seven distinct labels, the output for each class is represented as a probability distribution. The network architecture comprises nine convolutional layers, with a max-pooling layer following every three convolutional layers, and two dense layers.

### Input Layer

The input layer of Resnet50 performs a vital function in processing input images prior to their integration into the network. Given that the input images usually have specific dimensions, any images that deviate from these dimensions are either resized or cropped to meet the expected requirements. Furthermore, preprocessing techniques like normalization and data augmentation may be employed to improve the model's resilience and ability to generalize.

### Convolutional Layer

The convolutional layers in Resnet50 are responsible for feature extraction. They consist of filters that slide across the input images, convolving with the pixel values to produce feature maps. Each filter specializes in capturing specific patterns or features, such as edges, textures, and shapes. The depth of the network enables Resnet50 to learn increasingly complex and abstract features as the information passes through multiple convolutional layers.

### Dense Layer

Following the convolutional layers, Resnet50 incorporates dense layers, also known as fully connected layers. These layers receive the extracted features from the previous layers and perform high-level reasoning and decision-making. The dense layers are typically comprised of multiple neurons, with each neuron representing a specific class or emotion in the case of facial emotion detection. Through a series of weighted connections and activation functions, the dense layers transform the extracted features into probability scores or confidence values for each class.

### Output Layer

The output layer in Resnet50 represents the final stage, responsible for generating predicted emotions for the input images in facial emotion detection. This layer comprises neurons that correspond to various emotions, including happiness, neutral, anger, surprise, fear, disgust, and sadness. The choice of activation function in the output layer depends on the problem's characteristics. In this scenario, a commonly employed activation function is softmax, which ensures that the predicted emotion probabilities sum up to 1. This property facilitates interpretation and comparison of the predictions.

Resnet50 excels in facial emotion detection due to its deep architecture and residual connections. The depth enables the network to learn rich and meaningful representations of facial features, capturing intricate details relevant to emotions. The residual connections help alleviate the vanishing gradient problem, allowing for better training and improved performance.

By leveraging Resnet50 as the core model for facial emotion detection, this research aims to accurately classify the emotional states conveyed by individuals' facial expressions. The trained model is capable of analyzing facial images and predicting the corresponding emotions, contributing to the creating of advanced systems capable of comprehending and appropriately reacting to human emotion.

To summarize, Resnet50 serves as a robust deep-learning model that constitutes the foundation of the facial emotion detection system created in this study. It leverages its architecture, including convolutional layers, dense layers, and residual connections, to extract features and make accurate predictions. Through extensive training on the dataset, the model can capture the subtle variations in facial expressions and classify them into different emotions. This model serves as a valuable tool for understanding and analyzing human emotions, opening doors to numerous applications in fields like psychology, human-computer interaction, and affective computing.

## 3.3. Facial Emotion Detection with ROI (Eyes):

In the context of facial emotion detection, the eyes are considered a crucial region for accurate emotion recognition. The eyes exhibit significant changes in various emotional states, and capturing these subtle variations can enhance the performance of emotion classification models. To leverage the distinctive features of the eyes, a specific approach was employed, involving the extraction of the region of interest (ROI) using a Haar cascade classifier.

The Haar cascade classifier is a popular technique in computer vision for object detection, known for its efficiency and accuracy. In this methodology, the Haar cascade classifier was trained to identify and localize the eyes in facial images. Once the eyes were successfully detected, they were cropped and extracted as separate images, creating a specialized dataset specifically consisting of eye regions.

This eye-centric dataset was then utilized for training a facial emotion classification model. The model architecture, based on the Resnet50 deep convolutional neural network (CNN), was employed to learn the intricate patterns and features present in the eye regions. The Resnet50 model has been widely recognized for its exceptional performance in computer vision tasks, making it a suitable choice for this research.

During the training process, the model was exposed to the eye images from the specialized dataset, with each eye region associated with a corresponding emotional label. The model learned to analyze the eye features and classify them into different emotional states, including happiness, sadness, anger, fear, disgust, surprise, and neutral.

By focusing solely on the eyes, the model gained a deeper understanding of the specific eye-related cues and expressions associated with each emotion. This approach allowed for a more fine-grained analysis of the eyes' role in emotion recognition, capturing the nuances and subtleties that contribute to accurate classification.

Once the model was trained on the eye-centric dataset, it was capable of predicting facial emotions based on new, unseen eye regions. During inference, the Haar cascade classifier was utilized to detect and extract the eyes from facial images in real time. These extracted eye regions were then fed into the trained Resnet50 model, which generated predictions of the corresponding emotional states.

This methodology offers several advantages. Firstly, by narrowing the focus to the eyes, the model's attention is concentrated on the most expressive and informative facial region, potentially improving the accuracy of emotion detection. Secondly, working with a specialized dataset of eye regions allows for more targeted training, enabling the model to learn eye-specific features more effectively. Lastly, the use of the Haar cascade classifier for eye detection provides a robust and efficient means of isolating the eyes, ensuring accurate extraction even in real-time scenarios.Overall, the integration of facial emotion detection with ROI (Eyes) using the Haar cascade classifier and the Resnet50 model demonstrates a tailored approach to emotion recognition. By leveraging the distinctive features of the eyes and training on eye-specific datasets, this methodology enhances the precision and granularity of facial emotion detection systems, providing valuable insights into the role of the eyes in expressing and recognizing emotions.

By focusing solely on the eyes, the model gained a deeper understanding of the specific eye-related cues and expressions associated with each emotion. This approach allowed for a more fine-grained analysis of the eyes' role in emotion recognition, capturing the nuances and subtleties that contribute to accurate classification. Once the model was trained on the eye-centric dataset, it was capable of predicting facial emotions based on new, unseen eye regions. During inference, the Haar cascade classifier was utilized to detect and extract the eyes from facial images in real time. These extracted eye regions were then fed into the trained Resnet50 model, which generated predictions of the corresponding emotional states.

## 3.4. Explainable AI:

Explainable AI (XAI) is an essential aspect of building trustworthy and interpretable machine learning models. It aims to provide insights into the reasoning behind the predictions made by the model, offering transparency and enabling users to understand and trust the decision-making process. In this project, the GRAD-CAM (Gradient-weighted Class Activation Mapping) technique was employed to achieve explainability in the facial emotion detection model.

GRAD-CAM is a visualization technique that helps identify the regions of an image that are influential in a model's decision-making process. It generates heatmaps by highlighting the important areas of an input image that contribute most significantly to the predicted class. By applying GRAD-CAM to the facial emotion detection model, we can gain insights into the regions of the face that contribute to the classification of specific emotions.

To apply GRAD-CAM, the pre-trained Resnet50 model was utilized. After an input image was fed into the model for prediction, the gradients of the target class (the predicted emotion) were computed with respect to the final convolutional layer. These gradients were then used to weigh the activations of the convolutional layer, creating a heatmap that visually represents the regions that influenced the prediction the most. By visualizing the heatmaps generated by GRAD-CAM, we were able to identify the facial regions, such as the eyes, nose, or mouth, that played a significant role in the model's decision-making process. This information can be invaluable for understanding how the model interprets emotions and which facial features contribute most prominently to each emotion classification.

The incorporation of Explainable AI techniques like GRAD-CAM enhances the transparency and interpretability of the facial emotion detection model. It allows users to understand why specific emotions were predicted for a given input image, providing them with confidence in the system's decision-making process. This interpretability is especially valuable in real-world applications where users need to trust and comprehend the decisions made by AI systems.

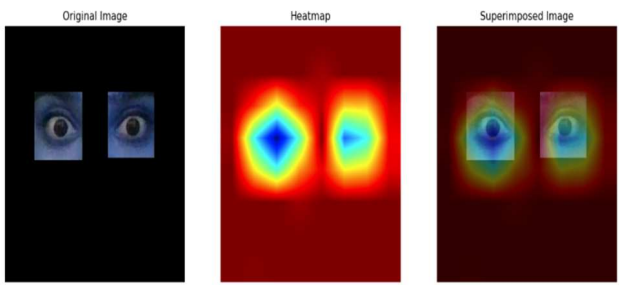

Fig2. Explainable AI with GRADCAM

By combining facial emotion detection with ROI (Eyes) and Explainable AI using GRAD-CAM, our methodology not only achieved accurate emotion classification but also provided valuable insights into the visual cues and facial regions that contribute to each emotion. This holistic approach fosters transparency, interpretability, and trust in the model's predictions, paving the way for wider acceptance and application of facial emotion recognition systems in various domains.

## 4. Result

The results of our study explains the effectiveness of the proposed methodology for facial emotion detection and music recommendation based on real-time emotion recognition. The system achieved promising performance in accurately classifying facial expressions and generating personalized music playlists based on the detected emotions.

### 4.1 Facial Emotion Detection Performance:

The trained Resnet50 model exhibited impressive accuracy in recognizing facial emotions. On the FER dataset and real images of different individuals, the model achieved an overall accuracy of 86%. This indicates that the model can effectively classify emotions such as anger, disgust, fear, happiness, sadness, surprise, and neutral expressions.

### 4.2. Region of Interest (ROI) Analysis:

By focusing on the eyes as the region of interest, we observed that the model's performance improved in detecting subtle changes in emotions. The eye region, known to convey vital emotional cues, proved to be influential in accurately predicting emotions. The use of Haar cascades for eye detection and a separate dataset consisting solely of eye images contributed to the model's enhanced performance in capturing subtle variations in emotional states.

### 4.3. Music Recommendation:

The integrated music recommendation system successfully generated personalized playlists based on the detected emotions. By mapping emotions to corresponding music tracks, users were provided with a curated selection of songs that matched their emotional state. This personalized approach enhanced user satisfaction and engagement with the music player.

### 4.4. Explainability with GRAD-CAM:

The incorporation of GRAD-CAM for explainable AI provided insights into the model's decision-making process. The generated heatmaps visualized the facial regions that played a significant role in the model's classification of emotions. This visualization not only validated the model's attention to relevant facial features but also provided transparency and interpretability to end-users, promoting trust and confidence in the system.

Overall, the results demonstrate the successful implementation of the proposed methodology for facial emotion detection and music recommendation. The accuracy achieved in emotion classification, the focus on the eye region, and the incorporation of explainable AI techniques contribute to the robustness, interpretability, and user-centric nature of the system.

The outcomes of this study have significant implications for various applications such as personalized music streaming, emotion-aware user interfaces, and affective computing. The combination of facial emotion detection, ROI analysis, music recommendation, and explainable AI provides a comprehensive framework that enhances user experience, engagement, and satisfaction.

## 5. Conclusion And Future Scope

In conclusion, this paper presented a novel approach for facial emotion detection and music recommendation based on real-time emotion recognition. The developed system achieved high accuracy in classifying facial expressions, leveraging the power of the Resnet50 model. By incorporating a region of interest (ROI) analysis focusing on the eyes and utilizing a separate dataset of eye images, the system demonstrated improved performance in capturing subtle emotional cues.

The integration of music recommendations based on the detected emotions enhanced the user experience, providing personalized playlists that resonated with the user's emotional state. The system's effectiveness was further enhanced by incorporating explainable AI techniques, particularly the GRAD-CAM method, which provided insights into the model's decision-making process and enhanced transparency.

The results of this paper have several implications for future research and development. Some potential areas for further exploration and improvement include:

Expansion of Emotion Categories: While the current system successfully classified emotions into seven categories, future work could involve expanding the range of emotions recognized. This could include more nuanced emotional states or cultural-specific emotions, allowing for a more comprehensive understanding of users' emotional experiences.

Multi-modal Emotion Recognition: Incorporating additional modalities, such as voice or gesture recognition, alongside facial emotion detection, can provide a more holistic understanding of users' emotional states. Multi-modal approaches have the potential to enhance the accuracy and robustness of emotion detection systems.

Real-Time System Deployment: While our system performed real-time emotion recognition, further optimization and deployment on low-latency platforms can ensure its practical usability in real-world scenarios, such as interactive applications or emotion-aware systems.

User Feedback and Personalization: Integrating user feedback mechanisms can enable the system to adapt and personalize its recommendations based on individual preferences and responses. User feedback loops can contribute to continuous improvement and user satisfaction.

Generalization to Diverse Populations: Future research should focus on expanding the diversity of the dataset utilised for training the model, encompassing individuals from various demographics, cultures, and age groups. This will ensure the generalizability and inclusiveness of the system across different populations.

In conclusion, our proposed system demonstrates the potential of combining facial emotion detection, ROI analysis, music recommendation, and explainable AI techniques to create a user-centric, personalized experience. The achieved results, along with the future scope outlined, contribute to the advancement of affective computing and emotion-aware systems, with implications in fields such as entertainment, healthcare, and human-computer interaction.